\documentclass[preprint,review,12pt]{elsarticle}

\usepackage{amssymb}

\usepackage{graphicx}
\usepackage{multirow}
\usepackage{booktabs}
\usepackage{tikz}
\usepackage{comment} 
\usepackage{amsmath,amssymb} 
\usepackage{color}
\usepackage{caption}
\usepackage{hyperref}

\newcommand{\etal}{\textit{et al}.~}

\newcommand{\ieno}{\textit{i}.\textit{e}.}

\newcommand{\egno}{\textit{e}.\textit{g}.} 
\newcommand{\etc}{\textit{etc}.}

\journal{Pattern Recognition}

\begin{document}

\begin{frontmatter}



\title{Structure-preserving feature alignment 
	for old photo colorization}


\affiliation[inst1]{organization={Department of Electronic Engineer and Information Science},
	addressline={University of Science and Technology of China}, 
	country={China}}

\tnotetext[label1]{Corresponding author}

\author[inst1]{Yingxue Pang}
\ead{pangyx@mail.ustc.edu.cn}

\author[inst1]{Xin Jin}
\ead{jinxustc@mail.ustc.edu.cn}

\author[inst1]{Jun Fu}
\ead{fujun@mail.ustc.edu.cn}

\author[inst1]{Zhibo Chen\corref{label1}}
\ead{chenzhibo@ustc.edu.cn}

\begin{abstract}
Deep learning techniques have made significant advancements in reference-based colorization by training on large-scale datasets. However, directly applying these methods to the task of colorizing old photos is challenging due to the lack of ground truth and the notorious domain gap between natural gray images and old photos. To address this issue, we propose a novel CNN-based algorithm called SFAC, \ieno, \textbf{S}tructure-preserving \textbf{F}eature \textbf{A}lignment \textbf{C}olorizer. SFAC is trained on only two images for old photo colorization, eliminating the reliance on big data and allowing direct processing of the old photo itself to overcome the domain gap problem. Our primary objective is to establish semantic correspondence between the two images, ensuring that semantically related objects have similar colors. We achieve this through a feature distribution alignment loss that remains robust to different metric choices.
However, utilizing robust semantic correspondence to transfer color from the reference to the old photo can result in inevitable structure distortions. To mitigate this, we introduce a structure-preserving mechanism that incorporates a perceptual constraint at the feature level and a frozen-updated pyramid at the pixel level. Extensive experiments demonstrate the effectiveness of our method for old photo colorization, as confirmed by qualitative and quantitative metrics.

\end{abstract}


\begin{highlights}
\item This paper proposes a novel CNN-based algorithm to address reference-based old photo colorization on only two images.
\item The proposed model, SFAC, is implemented as a multi-scale framework considering the data limitation. In each scale, we regard color transfer as a \textit{feature distribution alignment} problem and we design a \textit{structure-preserving mechanism} to avoid content distortions in the feature level and pixel level.

\item Extensive experiments on multiple benchmarks demonstrate the effectiveness and superiority of our method for old photo colorization in terms of qualitative and quantitative metrics.
\end{highlights}


\end{frontmatter}


\section{Introduction}

\begin{figure}[ht]
	\centering
 	\includegraphics[width=\textwidth]{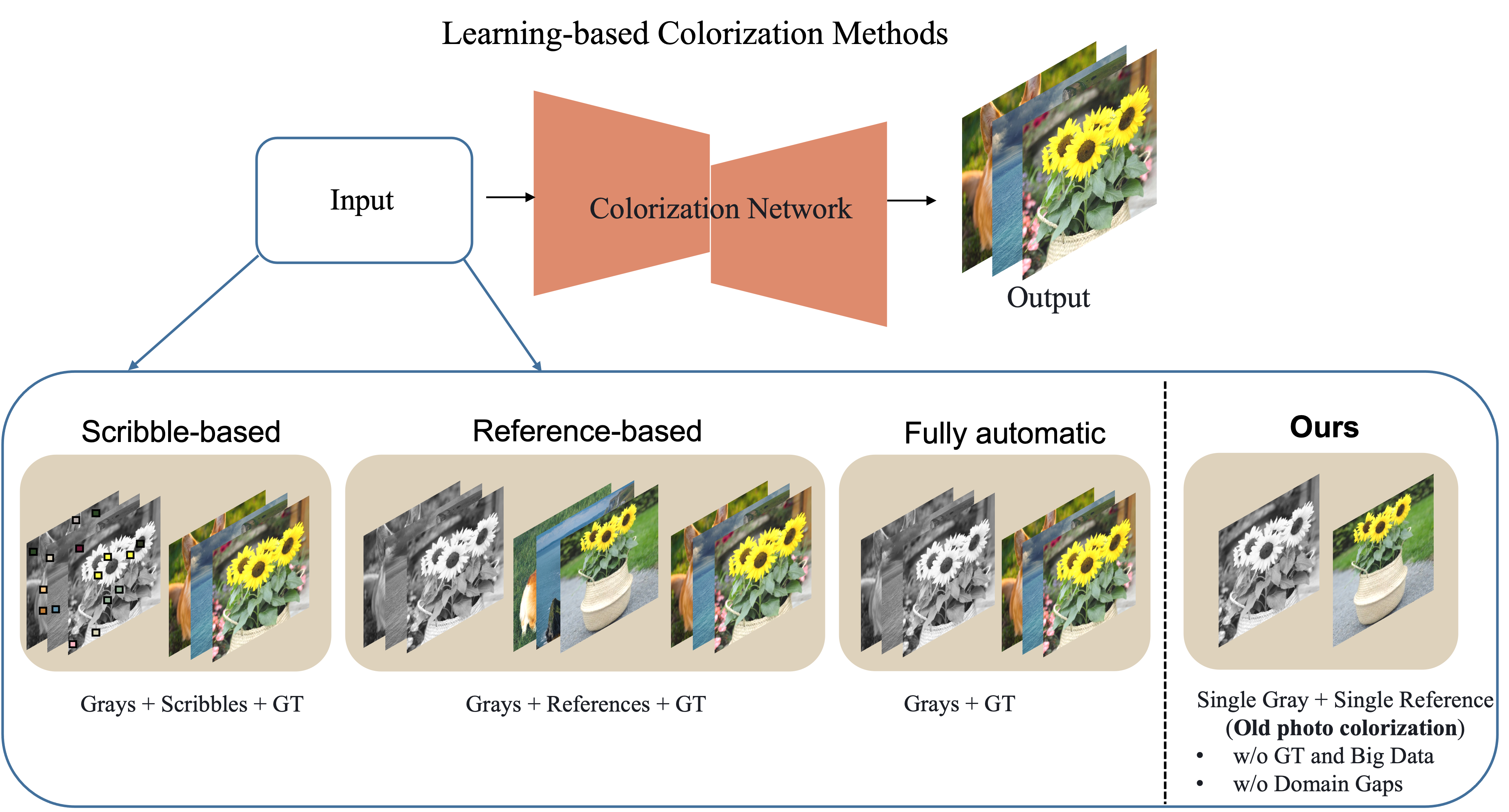}
	\caption{An overview of learning-based colorization methods. This figure shows the differences between different methods from the perspective of the training data input to the colorization network. }
	\label{fig:intro1}
\end{figure}

Image colorization aims to add style to gray photos by attaching vivid colors. 
This task is challenging because a grayscale image can correspond to multiple plausible colorizations (\egno, a car could be colored red, green, or blue). Early attempts~\cite{levin2004colorization,noda2006colorization} rely on user interactions to propagate colors from predefined scribbles as seed pixels throughout the image. This method, however, is labor-intensive and unsuitable for inexperienced individuals lacking professional skills and artistic sensibilities. 
To alleviate user burdens, some studies~\cite{welsh2002,liu2012automatic} employ a color reference image instead of scribbles, matching colors based on simple SIFT or SURF features and often leading to unsatisfactory results.
Recently, learning-based methods~\cite{he2018deep,deoldify} have emerged as promising solutions for colorization, benefiting from the powerful modeling capabilities of deep learning.

Fig.~\ref{fig:intro1} illustrates the overview of existing learning-based methods as scribble-based, reference-based, and fully automated, depending on their color information sources. Despite their impressive performance, these methods rely on deep and intricately designed CNNs that require extensive training on large-scale natural images over several days or weeks. And such data-rich CNN-based methods are often impractical for old photo colorization due to two primary reasons: (1) \textbf{No ground truth}: It is laborious and time-consuming to collect sufficient colored old photos to construct training pairs that meet the data requirements of existing data-intensive CNN-based models. (2)\textbf{Domain gaps}: CNN-based models are well-trained with large natural images that cannot be quickly generalized to characteristic historical photographs, resulting in significant performance degradation. The ideal clean grayscale images obtained with a predetermined bleaching process (\egno, linear weighting of the R, G, B channels) exhibit quite different data distribution to old photos, which undergo uncontrollable fading and long-time improper storage with deteriorating distortions (\egno, scratches, noise, and dirt)\cite{wan2020bringing}.

Therefore, some CNN-based color transfer algorithms~\cite{GU2022108716,liao2017visual,he2019progressive} trained with only two images are brought to our attention. When applying one old photo and one reference image, they can naturally extricate from the dependence on big data and avoid the domain gap by directly colorizing the old photo itself. 
However, Gu~\etal~\cite{GU2022108716} fail to leverage the rich prior contained in deep features learned from pre-trained CNNs, as their proposed Gaussian mixture model does not incorporate this.
Liao~\etal~\cite{liao2017visual} propose a two-step framework that utilizes deep features to establish semantic correspondence and subsequently replaces source pixels with corresponding reference pixels, known as local color transfer. However, this approach is susceptible to color discrepancies and may introduce local structural distortions due to the inseparable encoding information of high-level semantic and low-level color within deep features. 
To address this, He~\etal~\cite{he2019progressive} design a novel local color transfer algorithm that combines linear transformation and edge-preserving filtering. They jointly optimize the semantic correspondence in the feature domain and the local color transfer in the pixel domain.
Nonetheless, this method has two limitations. Firstly, finding semantic correspondence in the feature domain using nearest neighbor field computation is a cumbersome process involving bidirectional randomized sampling, propagation, and bidirectional similarity voting. Secondly, linear transformation and edge-preserving filtering only in the pixel domain can alleviate the inseparable problem, but cannot solve it due to the inherent traits of CNN-based deep features.

\begin{figure}[ht]
	\centering
	\includegraphics[width=\textwidth]{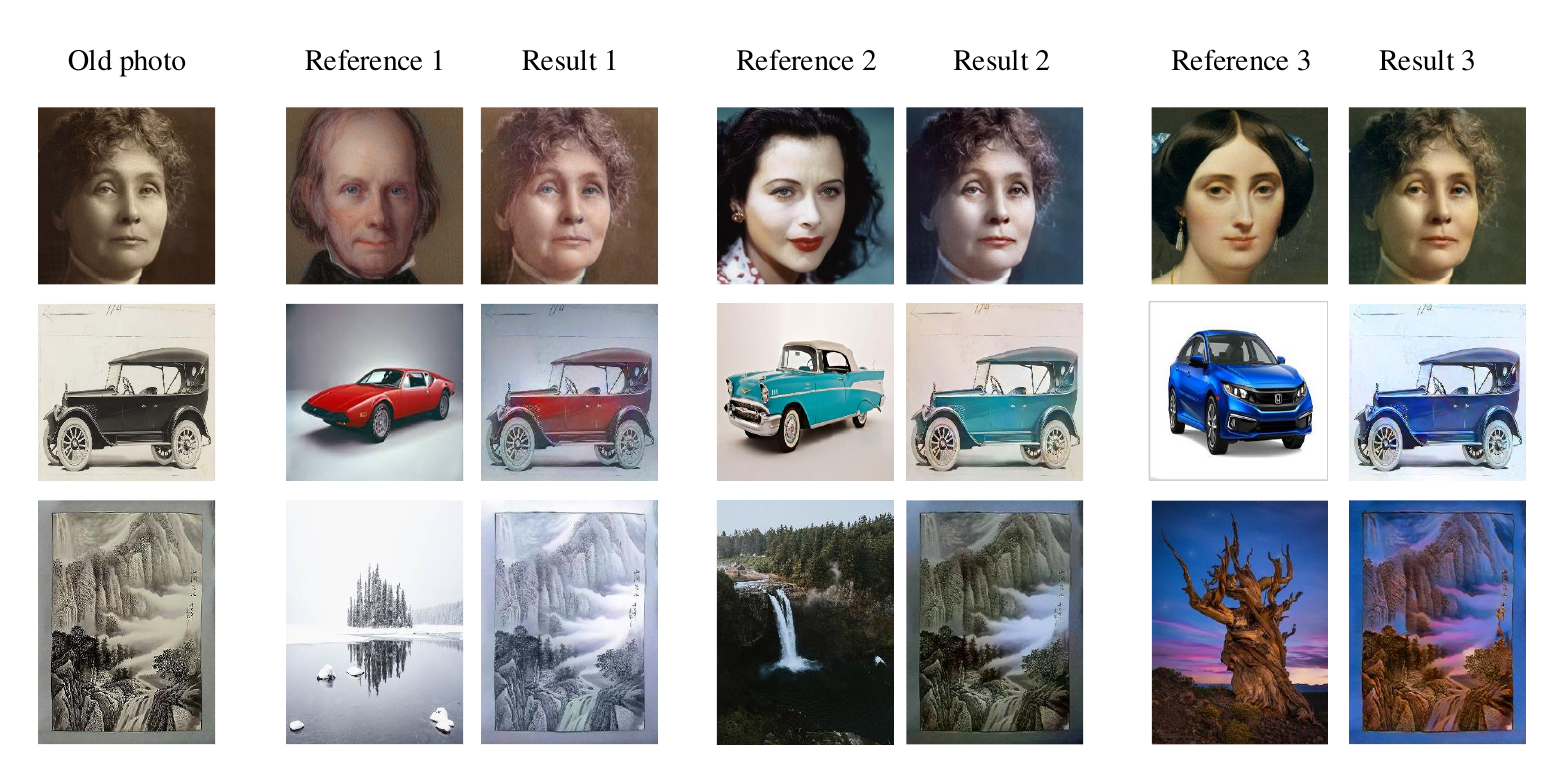}
	\caption{Our SFAC achieves reference-based old photo colorization with only two images and produces multiple plausible results when taking different color images as references. }
	\label{fig:intro}
\end{figure}

Instead, we propose a novel CNN-based algorithm trained on only two images for old photo colorization, \ieno, \textbf{S}tructure-preserving \textbf{F}eature \textbf{A}lignment \textbf{C}olorizer (SFAC). First, we greatly simplified the previously cumbersome process by designing a robust feature distribution alignment loss to implicitly establish the semantic correspondence. Second, we propose to preserve the structures in both the feature and pixel domains by transformer-based deep features to effectively ignore the effects of color and a frozen-updated Laplacian pyramid to further keep the fine structures. As shown in Fig.~\ref{fig:intro}, our SFAC successfully generates a variety of plausible results when taking different color images as references.  And more qualitative results are included in the supplemental material.

The main contributions of this paper are summarized as follows:
\begin{itemize}
	\item We propose a novel CNN-based algorithm to address reference-based old photo colorization on only two images.
	\item The proposed model, SFAC, is implemented as a multi-scale framework considering the data limitation. In each scale, we regard color transfer as a \textit{feature distribution alignment} problem and we design a \textit{structure-preserving mechanism} to avoid content distortions in the feature level and pixel level.
	
	\item Extensive experiments on multiple benchmarks demonstrate the effectiveness and superiority of our method for old photo colorization in terms of qualitative and quantitative metrics.

\end{itemize}

\section{Related Work}
\subsection{Colorization Approaches}

\subsubsection{Scribble-based Colorization Approaches}
These methods~\cite{levin2004colorization,noda2006colorization} need the user to draw color strokes over the grayscale image, then diffuse or propagate the colors from the strokes outward across the image. 
However, these algorithms are laborious, time-consuming, and require substantial manual effort. Users must provide extensive scribbles for various heterogeneous regions in grayscale images. In contrast, our method only requires a single color image as a reference to extract chrominance information and significantly reduces the burden on users.

\subsubsection{Reference-based Colorization Approaches}
Several attempts~\cite{welsh2002,liu2012automatic} have been made to transfer colors from a reference color image to a grayscale image without intensive manual efforts.  
However, they use pixel-level sparse color distribution representations (\egno, swatches, segments, superpixel, \etc) and could hardly capture more unified and higher-level color representations containing pixel and semantic information simultaneously.

More recent attention has focused on deep neural networks to capture color distribution automatically.
He~\etal\cite{10.1145/3197517.3201365} leverage deep image analogy~\cite{liao2017visual} to acquire a similarity map between gray images and reference images, then adopts an Unet to predict the chrominance channels of gray images. Zhang~\etal\cite{zhang2019deep} present a nonlocal network to predict the similarity map followed by an Unet as the second step. Xiao~\etal\cite{xiao2020example} define reference-based colorization as a multimodal classification problem.
Our SFAC is also a learning-based CNN approach but is trained on only two images without two well-designed neural networks and large-scale training data.

\subsubsection{Automatic Colorization Approaches}
The automatic colorization of grayscale images using deep learning techniques, without user intervention, has gained significant attention in recent years. Cheng~\etal\cite{cheng2015deep} pioneer the use of deep neural networks to address this problem by formulating it as a regression task. Recent studies~\cite{iizuka2016let,deoldify,vitoria2020chromagan} in fully automatic colorization have achieved remarkable success, benefiting from advancements in deep neural networks and the availability of large training datasets.
However, these methods rely on deterministic mapping and often struggle to generate images with the desired colors. In contrast, our SFAC approach offers the advantage of generating multiple plausible results by leveraging different color images as references.

\subsection{Feature Distribution Alignment}
Domain Adaptation (DA)~\cite{rahman2020correlation} aims to leverage the abundant labeled training samples in the source domain to facilitate learning in a new target domain by reducing the distribution divergency. 
Feature distribution alignment (FDA) is one effective way to settle DA by minimizing the discrepancy between the re-weighted source and target distributions to deal with domain shift~\cite{GAO2022108616,kiran2022incremental} or by minimizing the discrepancy between the marginal and the conditional distributions to learn domain-invariant feature representation~\cite{WANG2022108595,XU2022108700}.
In this paper, we utilize FDA to minimize the divergence of feature distributions with color characteristics extracted from the colorful reference and the colorized result. 

\section{Method}
\subsection{Framework Overview}
\begin{figure}[ht]
	\centering
	\includegraphics[width=1\textwidth]{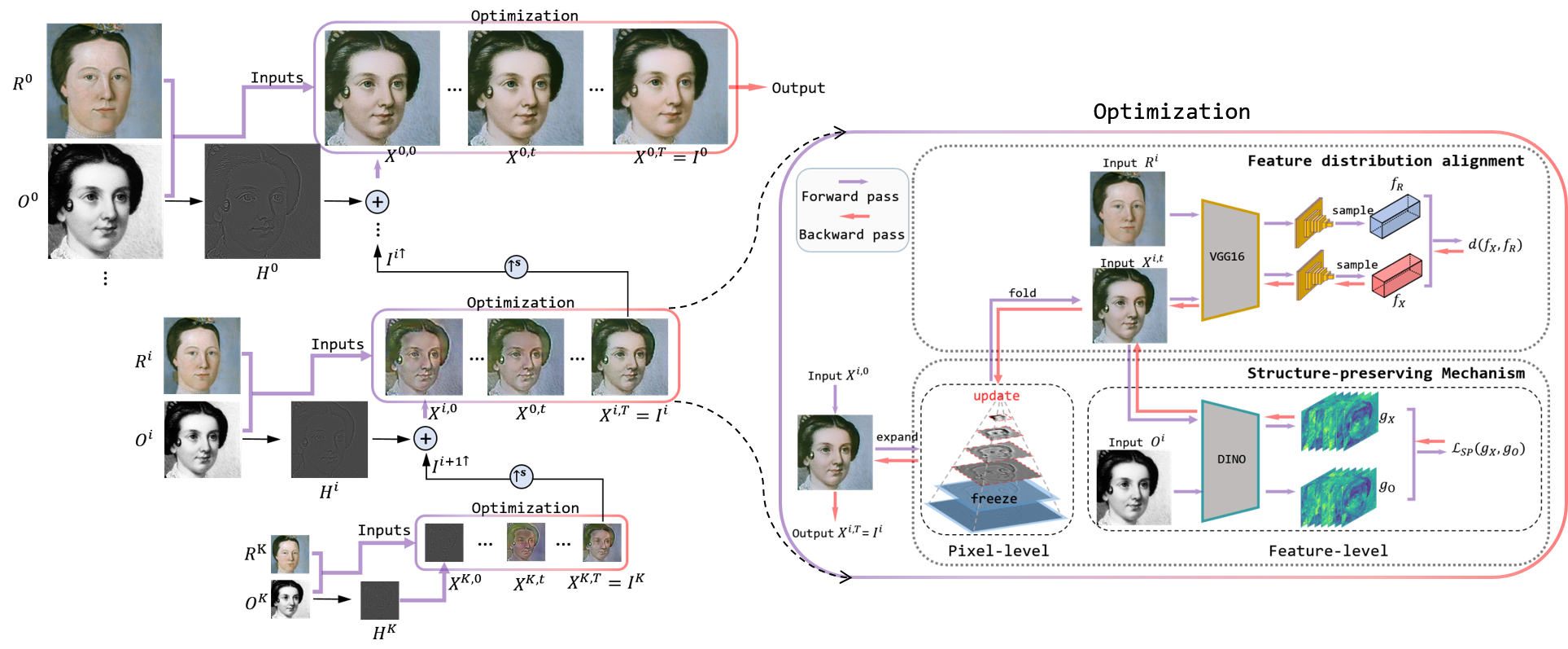}
	\caption{Our SFAC for reference-based old photo colorization. Left: The entire multi-scale framework of SFAC, which takes an old image $O^{i}$, a color reference $R^{i}$ and an initial result $X^{i,0}$ as inputs to iteratively optimize the result $X^{i,t}$ in each scale. SFAC progressively generates the final output $I^{0}=X^{0,T}$ through multiple scales. Right: The detailed optimization process in each scale including the FDA module (top) to transfer color and the SPM module (bottom) to maintain structure.}
	\label{fig:framework}
\end{figure}

Our goal is to colorize one grayscale old photo $\textbf{O}$ based on one color reference image $\textbf{R}$. As such, we build a \textbf{S}tructure-preserving \textbf{F}eature \textbf{A}lignment \textbf{C}olorizer (SFAC) in the RGB color space. Notably, SFAC focuses on the scenario that two inputs share related semantical labels, but may vary dramatically in content or structure as shown in Fig.~\ref{fig:intro}. Considering the limited training data, SFAC is implemented as a multi-scale framework to progressively transfer colors, and thus we can fully leverage the two images. 
Compared with processing two images directly on the finest scale, first coloring on the coarse scale and then linearly interpolating the result to the finer scale as a new initial solution yields faster convergence.

As shown in Fig.~\ref{fig:framework}~(Left), we adopt a hierarchical structure across multiple scales in a coarse-to-fine fashion. The old photo colorization task takes a grayscale old image, a color reference image, and an initial colorized result (upsampled from the previous scale) as inputs, and aims to estimate the color old image at this scale as 
\begin{equation}
	\textbf{I}^{i} = \textbf{Net}_{SR}(\textbf{O}^{i},\textbf{R}^{i},\textbf{I}^{i+1\uparrow}),
\end{equation}
where $i\in\{0,1,\dots,K\}$ is the scale index and $i = 0$ represents the finest scale. $\textbf{O}^{i}$, $\textbf{R}^{i}$, and $\textbf{I}^{i}$ are the grayscale old photo, color reference image, and the estimated color old photo at the $i$-th scale. $(\cdot)\uparrow$ denotes the upsampling operator. $\textbf{Net}_{SR}$ is the proposed scale-wise network, which adopts an optimization-based iterative process. As shown in Fig.~\ref{fig:framework} (Right), at the $i$-th scale, $\textbf{Net}_{SR}$ aims to optimize the output color old photo towards minimizing the following objective function:
\begin{equation}
	\mathcal{L} = \lambda_{\text{FDA}} \mathcal{L}_{\text{FDA}}(\textbf{X}^{i,t}, \textbf{R}^{i}) + \lambda_{\text{P}} \mathcal{L}_{\text{P}}(\textbf{X}^{i,t}, \textbf{O}^{i}),
\end{equation}
where $\lambda_{\text{FDA}}$ and $\lambda_{\text{P}}$ control the importance of the featue distribution alignment loss term $\mathcal{L}_{\text{FDA}}$ and the perceptual loss term $\mathcal{L}_{\text{P}}$. $\textbf{X}^{i,t}$ denotes the colorized old photo at the $t$-th iteration where $t\in\{0,1,\dots,T\}$. We will describe these two loss functions in detail in the following sections. Each update of the output color old photo follows the gradient,

\begin{equation}
	\begin{aligned}
		&\textbf{X}^{i,t+1} \leftarrow \textbf{X}^{i,t} - \gamma\frac{\partial \mathcal{L}}{\partial \textbf{X}^{i,t}},\quad\text{where}\\
		&\textbf{X}^{i,0}=
		\begin{cases}
			\textbf{I}^{i+1\uparrow} + \textbf{H}^{i},& \text{$i<K$;}\\
			\textbf{H}^{i},& \text{$i=K$.}
		\end{cases}
	\end{aligned}
\end{equation}
where $\gamma$ represents the learning rate, $K$ denotes the maximum number of scales, $\textbf{H}^{i}$ denotes the high-frequency part of $\textbf{O}^{i}$. $\textbf{X}^{i,T} = \textbf{I}^{i}$ is the final result of the scale $i$, where $T$ is the maximum number of iterations. Next, we will detail feature distribution alignment and structure-preserving mechanism in sequence. 
For conciseness, we omit the scale index $i$ and the iteration index $t$ in the following sections. 

\subsection{Feature Distribution Alignment (FDA)}
The goal of reference-based image colorization is to ensure that semantically related objects in the reference image and the old photo have similar colors. Therefore, the first and most important issue is how to leverage the established semantic correspondence to transfer colors between two images. Previous methods first build correspondence by a subnet~\cite{he2018deep,zhang2019deep} or NNF computation~\cite{liao2017visual,he2019progressive}, and then use it to guide color transfer by another subnet or linear transformation. Differently, we simplify the above two-stage process and directly transfer color by aligning the feature distributions of the colorized result and the reference image via a feature distribution alignment (FDA) loss, where the semantic correspondence is built indirectly through the features extracted from a VGG16 recognition network pre-trained on ImageNet. 

Specifically, pre-trained CNNs could effectively extract semantic and appearance information from arbitrary images by different positions and numbers of layers~\cite{gatys2016image,huang2017arbitrary}. In this way, the semantically similar parts of the two images should exhibit similar network responses, i.e., feature distributions. Let $f_{\textbf{X}} =\{\Phi(\textbf{X})_{l_1}\uparrow,...,\Phi(\textbf{X})_{l_N}\uparrow\}$ and $f_{\textbf{R}} =\{\Phi(\textbf{R})_{l_1}\uparrow,...,\Phi(\textbf{R})_{l_N}\uparrow\}$ mean the captured hierarchical representations of $\textbf{X}$ and $\textbf{R}$, where $\Phi(\cdot)_{l_1}$ denotes the feature maps of the $l_1$-th layer of the VGG16 network $\Phi$, and $(\cdot)\uparrow$ denotes upsample the input feature maps to the shape of $\textbf{X}$ with bilinear interpolation. Therefore, we achieve color transferring by minimizing the feature discrepancy between the colorized result and the reference image,
\begin{equation}
	\underset{\textbf{X}}{\textbf{argmin}} \qquad d(f_{\textbf{X}}, f_{\textbf{R}}),
\end{equation}
where $d(\cdot)$ denotes the distribution distance metric.

A variety of distribution distance metrics can be incorporated into the FDA system to minimize the discrepancy, and our experiments, presented in Fig~\ref{fig:FDA}, show that all these metrics can achieve the success of old photo colorization even if they do not perform the same results. Here we introduce four representative metrics, including Feature Statistics (FS), Contextual Loss (CX), Central Moment Discrepancy (CMD), and Relaxed Earth Mover's Distance (rEMD), and experimentally study their influence w.r.t the final performance. Our method is robust to this metric choice:

\noindent{\bf Feature Statistics (FS).} Numerous style transfer works~\cite{gatys2016image,huang2017arbitrary} have revealed that the feature statistics in convolutional layers of a deep neural network encode the style information of an image. Since chrominance information is the most important trait of style, we use FS as a metric in our FDA to match the second-order feature statistics:
	\begin{equation}
		d(f_{\textbf{X}}, f_{\textbf{R}})= ||\mu_{f_{\textbf{X}}}-\mu_{f_{\textbf{R}}}||_{2} + ||\sigma_{f_{\textbf{X}}}-\sigma_{f_{\textbf{R}}}||_{2}, 
	\end{equation}
	where $\mu_{f_{\textbf{X}}}$ and $\sigma_{f_{\textbf{X}}}$ are the mean and variance of the feature distribution $f_{\textbf{X}}\in{\mathbb{R}^{hw}}$ extracted from $\textbf{X}$ in each channel across both spatial dimensions,
	\begin{align}                                 
		&\mu_{f_{\textbf{X}}} = \frac{1}{\text{hw}}\sum_{h}\sum_{w}f_{\textbf{X}},\nonumber\\
		&\sigma^{2}_{f_{\textbf{X}}}= \frac{1}{\text{hw}}\sum_{h}\sum_{w}(f_{\textbf{X}}-\mu_{f_{\textbf{X}}})^{2}+\epsilon.
	\end{align}
	$\epsilon=1e-5$ is a fixed value, and $\mu_{f_{\textbf{R}}}$ and $\sigma_{f_{\textbf{R}}}$ correspond to the feature distribution $f_{\textbf{R}}\in{\mathbb{R}^{hw}}$.

\noindent{\bf Contextual Loss (CX).} CX loss~\cite{mechrez2018contextual} considers the feature similarities instead of their spatial locations, therefore it is suitable to our color transfer task to alleviate the structure distortion during FDA and avoid aligning content statistics. Mathematically,
\begin{equation}
	d(f_{\textbf{X}}, f_{\textbf{R}})=-log(\text{CX}(f_{\textbf{X}},f_{\textbf{R}})),
\end{equation}
where $\text{CX}$ is defined as the contextual similarity between two feature distributions $f_{\textbf{X}}=\{f_{\textbf{X}}^{i}\}$ and $f_{\textbf{R}}=\{f_{\textbf{R}}^{j}\}$ after sampling $N$ points from them respectively,
\begin{align}
	& \text{CX}(f_{\textbf{X}},f_{\textbf{R}})= \frac{1}{N}\sum_{j}\mathop{max}\limits_{i}A_{ij},\nonumber\\
	& A_{ij} = \frac{exp(1-\widetilde{d}_{ij}/h)}{\sum_{k}exp(1-\widetilde{d}_{ik}/h)}.
\end{align}
$h>0$ is a band-width parameter and $\widetilde{d}_{ij}$ denotes the normalized cosine distance between $f_{\textbf{X}}^{i}$ and $f_{\textbf{R}}^{j}$.

\noindent{\bf Central Moment Discrepancy (CMD).} CMD~\cite{ZELLINGER2019174} aims to learn a domain-invariant representation for adaptation by utilizing moment sequences as an equivalent representation of probability distributions, which also extends the moment match to higher orders without any computationally expensive distance- and kernel matrix computations, leading to a more compact distance measure. We use it to measure the distance of two feature distributions:
\begin{equation}
	d(f_{\textbf{X}}, f_{\textbf{R}})= \text{CMD}(f_{\textbf{X}}, f_{\textbf{R}}),
\end{equation}
	where $\text{CMD}$ is defined as follows~\cite{ZELLINGER2019174},
	\begin{equation}
		\begin{aligned}
			&\text{CMD}(f_{\textbf{X}}, f_{\textbf{R}}) = \sum_{i=1}^{k}\alpha_{i}{||c_{i}(f_{\textbf{X}})-c_{i}(f_{\textbf{R}})||}_{2},\\
			&c_{i}(f_{\textbf{X}})=
			\begin{cases}
				\mathbb{E}[f_{\textbf{X}}],& \text{$i=1$;}\\
				\mathbb{E}[\nu^{i}({f_{\textbf{X}}-\mathbb{E}[f_{\textbf{X}}])}],& \text{$i\ge{2}$.}
			\end{cases}\\
			&\nu^{i}: \mathbb{R}^{m}\to{\mathbb{R}^{\frac{(i+1)^{m-1}}{(m-1)^{!}}}}.
		\end{aligned}
	\end{equation}
	$\alpha_{i}\ge{0}$, $\mathbb{E}[f_{\textbf{X}}]$ is the expectation of $f_{\textbf{X}}$ and $\nu^{i}$ is the vector-valued function mapping a $m$-dimension vector $f_{\textbf{X}}$ to its $\frac{(i+1)^{m-1}}{(m-1)^{!}}$ monomial values of order $i$.

%
%

\noindent{\bf Relaxed Earth Mover's Distance (rEMD).} EMD~\cite{zhu2014sparse} is introduced to measure the distance as a minimum cumulative cost of transforming knowledge from one distribution to another, and the desired transformation should make the two distributions close.  Here we use a more relaxed version, \ieno, rEMD~\cite{rabin2014adaptive} to accelerate the process of feature distribution alignment: 
	\begin{equation}
		d(f_{\textbf{X}},f_{\textbf{R}})=max\{\text{rEMD}(f_{\textbf{X}}, f_{\textbf{R}}),\text{rEMD}(f_{\textbf{R}}, f_{\textbf{X}})\},
	\end{equation}
	where $\text{rEMD}(f_{\textbf{X}},f_{\textbf{R}})$ is defined as follows,
	\begin{equation}
		\begin{aligned}
			&\text{rEMD}(f_{\textbf{X}},f_{\textbf{R}}) = \mathop{argmin}\limits_{T\in\mathbb{R}_{+}}\sum_{ij}T_{ij}M_{ij},\\
			&s.t.\quad T^{\top}\mathbf{1}_{f_{\textbf{X}}}=\mu_{f_{\textbf{R}}}.
		\end{aligned}
	\end{equation}
	$\mu_{f_{\textbf{R}}}$ is the discrete empirical distribution of $f_{\textbf{R}}$, $T_{ij}$ denotes the optimal amount of mass to move from $f_{\textbf{X}}$ to $f_{\textbf{R}}$ to obtain an overall minimum cost and $M_{ij}$ is the cost matrix (cosine similarity matrix) to represent the distance between $f_{\textbf{X}}$ and $f_{\textbf{R}}$.

\subsection{Structure-Preserving Mechanism (SPM)}
Nevertheless, optimizing FDA with only two images inevitably results in content distortion since the univocal constraint would force the colored result to reach unanimity with the reference, whether color or structure, seen in Fig.~\ref{fig:ab1} (with FDA only). To alleviate this issue, we further propose a structure-preserving mechanism including feature and pixel level constraints.

\begin{figure}[ht]
	\centering
	\includegraphics[width=\textwidth]{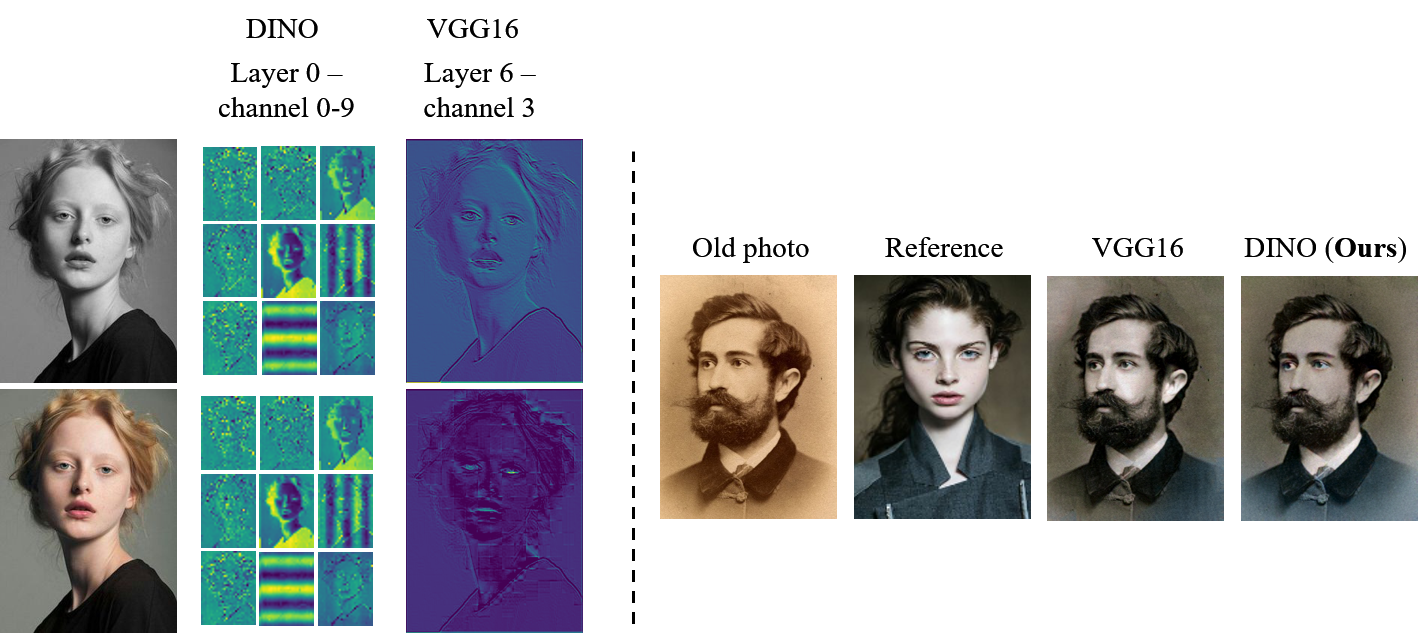}
	\caption{(Left) The visualization of feature maps extracted from DINO and VGG16 respectively according to one gray image and its colorful ground truth. The feature maps of grayscale and color images extracted by VGG16 show great differences in the same channel of the same layer, while DINO acquires very similar results with a tiny gap.}
	\label{fig:dino}
\end{figure}

\subsubsection{Feature Level (SPM-F)} 
To maintain the raw structure consistency between the output \textbf{X} and the old photo \textbf{O} in the feature space, it is necessary to employ a spatially invariant representation that accurately represents the scene structure while disregarding color values. Previous methods~\cite{liao2017visual,he2019progressive} utilize pre-trained CNNs to extract features with structural information. However, these extracted features capture both structural and color information, resulting in significant color inhomogeneity, as evident in Fig.~\ref{fig:dino}, where VGG16 generates different feature distributions for the gray image and the color image. This occurs because the structural features extracted by VGG16 contain inseparable color information, impeding color transfer and structure preservation. Consequently, using VGG16 directly to extract structural representations for colorization is infeasible. Fortunately, we experimentally find that the latest vision transformer DINO~\cite{caron2021emerging} effectively captures scene structure while disregarding color effects. As shown in Fig.~\ref{fig:dino}, DINO exhibits minimal discrepancies in network responses between the two images and produces more desirable outputs, such as homogeneous color in clothing and background.
Therefore, we use the pre-trained DINO to respectively extract the feature maps $g_{\textbf{X}} =\{\Psi(\textbf{X})_{l_1},...,\Psi(\textbf{X})_{l_N}\}$ and $g_{\textbf{O}} =\{\Psi(\textbf{O})_{l_1},...,\Psi(\textbf{O})_{l_N}\}$, where $\Psi(\cdot)_{l_1}$ denotes the feature maps of the $l_1$-th layer of the DINO network $\Psi$. Then we compute the perceptual loss $\mathcal{L}_{\text{P}}$ by:
\begin{equation}
	\mathcal{L}_{\text{P}}(g_{\textbf{X}}, g_{\textbf{O}})=\frac{1}{l_N}\sum_{i=l_1}^{l_N}(1-\frac{\Psi(\textbf{X})_{i}\cdot\Psi(\textbf{O})_i}{||\Psi(\textbf{X})_i ||||\Psi(\textbf{O})_i||}).
\end{equation}

\label{sec:expr}

\noindent{\bf Pixel Level (SPM-P).}
SPM-F does not perform well in preserving the coherence of fine structures and details since feature representations inevitably lose some low-level information. Therefore, we impose an explicit restriction in the pixel level to keep the details and edges of the colorized result \textbf{X} unchanged after colorization. In detail, we utilize a Laplacian pyramid (LP)~\cite{burt1987laplacian} to linearly decompose \textbf{X} into a set of high- and low-pass bands using recursive computation, where the sum of resulting bands can perfectly reproduce the original image. The detailed process of decomposition is as follows. 

After getting the optimized result $\textbf{X}=\textbf{X}_{0}\in{\mathbb{R}^{H\times{W}}}$, LP first blurs and decimates it with a downsampling operator $down(\cdot)$ to get a new version $\textbf{X}_{1}\in{\mathbb{R}^{\frac{H}{2}\times{\frac{W}{2}}}}$. To allow a linear invertible reconstruction, LP records the high-frequency residual $H_{0}=\textbf{X}_{0}-\textbf{X}_{0}^{'}$ where $\textbf{X}_{0}^{'}\in{\mathbb{R}^{H\times{W}}}$ is the upsampled version of $\textbf{X}_{1}$ using an upsampling operator $up(\cdot)$. Overall,
\begin{equation}
	H_{0}= \textbf{X}_{0}-up(down(\textbf{X}_{0})).
	\label{eq:lap}
\end{equation}
Conducting $L$ times of the process mentioned above iteratively, we decompose the image $I_{0}$ into an LP that consists of a set of band-pass components $\mathcal{H}=[H_{0},\dots,H_{L-1}]$ and a low-frequency residual image $\textbf{X}_{L}$.

To preserve fine structures and boundary information, we freeze the first two high-frequency bands and only update the remaining bands of \textbf{X}'s LP during the iterative updating process, as illustrated in Fig~\ref{fig:framework} (Right: Pixel-level). This frozen-updated pyramid ensures explicit consistency of details and boundaries between the colorized result and the old photo.

\subsection{Implementation details}
We collect $400$ old photos and $4000$ reference images from Pexels\footnote{https://www.pexels.com/}, Flickr\footnote{https://www.flickr.com/}, MetFaces~\cite{Karras2020ada} containing portrait photography, architectural photography, landscape photography as well as high-quality human faces images collected from Metropolitan Museum of Art.

We train our model using the RM-Sprop optimizer with an initial learning rate $0.002$, and we decay the learning rate to 0.001 on the last scale. We set the number of scales $K=3$ and train iterations $T=200$ for each scale. And we set the layers of the Laplacian pyramid $L=5$. For FDA, we use the convolutional layers of VGG16 including layers 1, 3, 6, 8, 11, 13, 15, 22 and 29 to construct a multi-level hypercolumn to represent color statistics. And for SPM-F, we use all the 12 output layers of DINO~\cite{caron2021emerging} and load the weight of the pre-trained model ViT-S/8 considering the model performance and parameter numbers. Generally, the weight of perceptual loss $\mathcal{L}_{\text{P}}$ is set to $1$. As for the weight of FDA loss $\mathcal{L}_{\text{FDA}}$, we set $2$ for FS, $6$ for CX, $0.5$ for CMD, and $1$ for rEMD. Our model requires 70 seconds on a single 2080-Ti GPU with an image of $256\times256$ size.

\section{Experiments}

In this section, we validate the performance of our model by comparing the image quality quantitatively and qualitatively with various types of image colorization methods. Moreover, we investigate the different properties of our model to illustrate the role of each designed component. 

\subsection{Baselines}
We compare our model, SFAC, with three types of algorithms, including (1) reference-based colorization: Welsh~\cite{welsh2002}, He~\cite{10.1145/3197517.3201365}, Zhang~\cite{zhang2019deep} and Xiao~\cite{xiao2020example}; (2) fully automatic colorization: Iizuka~\cite{iizuka2016let}, Larsson~\cite{larsson2016learning}, DeOldify~\cite{deoldify} and ChromaGAN~\cite{vitoria2020chromagan};
(3) photorealistic style transfer: Liao~\cite{liao2017visual}, He~\cite{he2019progressive}, PhotoWCT~\cite{li2018closed} and WCT2~\cite{yoo2019photorealistic}. For all baseline methods, we directly use their released codes or demos with default configurations for a fair comparison.

\subsection{Evaluation metric}
\noindent(1) \textbf{Single Image Fr\'{e}chet Inception Distance (SIFID)  \cite{shaham2019singan}}: We leverage SIFID to evaluate the performance of color transfer. 
A lower SIFID score indicates that the old photo learns more chrominance information from the reference image.

\noindent(2) \textbf{Fr\'{e}chet Inception Distance (FID) \cite{heusel2017gans}}: We compute FID between a collection of colorized photos and reference images to evaluate the quality of colorized photos. A lower FID score means better performance.

\begin{figure}[!t]
	\centering
	\includegraphics[width=\textwidth]{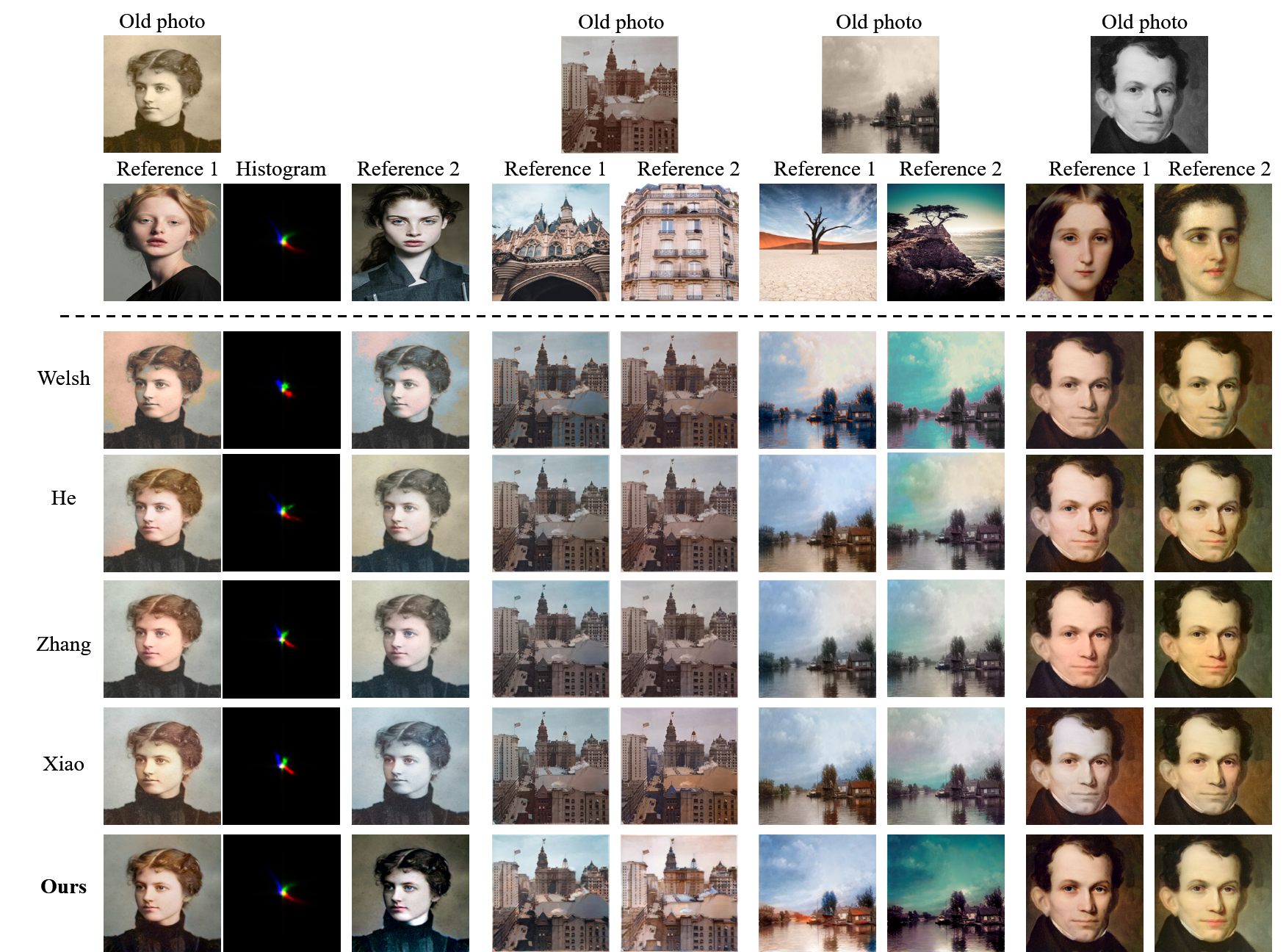}
	\caption{Comparisons with reference-based colorization methods: Welsh~\cite{welsh2002}, He~\cite{10.1145/3197517.3201365}, Zhang~\cite{zhang2019deep}, Xiao~\cite{xiao2020example}.}
	\label{fig:exper1}
\end{figure}

\noindent(3) \textbf{Color Histogram (CH) \cite{afifi2021histogan}}: It is necessary to quantify how well the model captures the colors of a reference image. We refer to Color Histogram (CH), as shown in Fig.~\ref{fig:exper1} (column 2), which represents the color distribution of an image while remaining decoupled from image-specific semantics, to compare the color histogram differences between the colorized old photo and corresponding reference image. We implement the Hellinger distance and KL divergence to measure the differences respectively. A lower CH score indicates that the colorized old photo shares more similar color statistics with its reference. 

\noindent(4) \textbf{User Study (US)}: To evaluate the overall performance of the colorization regarding both the photorealism and color transfer effects, we also perform three user studies based on human perception. More details are presented in the following sections.

\setlength{\tabcolsep}{4pt}

\subsection{Colorization Results Comparisons}
\noindent{\textbf{Comparison with Reference-based Colorization}.} Fig.~\ref{fig:exper1} presents the qualitative comparisons of the state of the art. We can see that Welsh~\etal\cite{welsh2002} often get stuck into local minimum and fail to explicitly enforce a contiguous assignment of colors. He~\etal\cite{10.1145/3197517.3201365}, Zhang~\etal\cite{zhang2019deep} and Xiao~\etal\cite{xiao2020example} suffer from severe model degradations due to domain gap, leading to noticeable color artifacts (\egno, column 1\&3) and unpleasant colorized results (\egno, column 6\&7). On the contrary, our model achieves more vivid and saturated colorized results. It transfers colors from reference as much as possible in a semantic-friendly way (\egno, column 6, 7 and eye-to-eye, mouth-to-mouth in the final column). For more comparison results, please refer to the supplemental material.

\begin{table}[ht]
	\centering
	\caption{SIFID, FID, CH and US scores across different reference-based colorization methods. The best scores are in bold.}
	\resizebox{\textwidth}{!}{
		\begin{tabular}{c|ccccccccccccccccc}
			\toprule
			\multirow{3}{*}{Configuration} & \multicolumn{4}{c}{\multirow{2}{*}{\begin{tabular}[c]{@{}c@{}}SIFID\\ (x10-4)\end{tabular}}} &  & FID &  & \multicolumn{8}{c}{CH} &  & US \\ \cline{7-7} \cline{9-16} \cline{18-18} 
			& \multicolumn{4}{c}{} &  & - &  & \multicolumn{4}{c|}{KL div.} & \multicolumn{4}{c}{H dis.} &  & - \\ \cline{2-5} \cline{9-16}
			& portrait & architectural & landscape & faces &  & - &  & portrait & architectural & landscape & \multicolumn{1}{c|}{faces} & portrait & architectural & landscape & faces &  & - \\ \cline{1-5} \cline{7-7} \cline{9-16} \cline{18-18} 
			Welsh~\cite{welsh2002} & 4.09 & 17.58 & 16.91 & 1.40 & & 250.76 & & 0.900 & 0.122 & 0.411 & 0.621 & 0.389 & 0.165 & 0.309 & 0.306 &  & 5.25\% \\
			He~\cite{10.1145/3197517.3201365} & 3.48 & 16.34 & 14.58 & 1.41 & & 231.02 & & 0.538 & 0.276 & 1.363 & 0.430 & 0.351 & 0.245 & 0.505 & 0.303 &  &   23.75\%\\
			Zhang~\cite{zhang2019deep} & 3.47 & 18.17 & 21.05 & 2.15 & & 243.99 & & 0.398 & 0.288 & 1.306 & 0.528 & 0.278 & 0.248 & 0.469 & 0.311 &  & 25.00\%  \\
			Xiao~\cite{xiao2020example} & 4.06 & 14.44 & 13.68 & 2.56 & & 245.71 & & 1.375 & 0.374 & 0.980 & 0.817 & 0.447 & 0.286 & 0.436 & 0.435 &  & 21.50\%\\
			\textbf{Ours} & \textbf{1.10} & \textbf{8.29} & \textbf{5.42} & \textbf{1.09} & & \textbf{224.05} & & \textbf{0.323} & \textbf{0.097} & \textbf{0.286} & \textbf{0.113} & \textbf{0.255} & \textbf{0.154} & \textbf{0.249} & \textbf{0.157} &  & -  \\
			\bottomrule
	\end{tabular}}
	\label{tab:ref}
\end{table}

To quantitatively evaluate each method, we conduct three objective comparisons regarding SIFID, FID and CH shown in the \textbf{SIFID}, \textbf{FID} and \textbf{CH} columns of Table.~\ref{tab:ref}. Our approach performs favorably against existing reference-based colorization methods in terms of these quantitative metrics. The lower SIFID and FID scores indicate that our results are more similar to the real data, or said higher image quality. Our CH score is also smaller than the baselines, which illustrates that the colorized result of our method shares the most similar color distributions to its corresponding reference. These results demonstrate our proposed method is more effective for old photo colorization. 

It is highly subjective to assess colorized results. Hence, a user study comparison is further designed
for the five approaches. For the user study (US), we randomly select 40 samples including the types of portrait photography, architectural photography, landscape photography, and artistic human faces. We display the old photo, reference image, and two colorized results from our model and another baseline method respectively on a webpage in random order. A group of $20$ participants unconnected with the project is asked to choose which one is more visually pleasing. Finally, we collect total $1600$ responses and $80$ responses for each comparison. We detail the preference percentage of each method in the \textbf{US} column of Table~\ref{tab:ref}, where most users prefer the colorized results generated by our model than Welsh~\cite{welsh2002}, He~\cite{10.1145/3197517.3201365}, Zhang~\cite{zhang2019deep}, and Xiao~\cite{xiao2020example}.

\begin{figure}[!t]
	\centering
	\includegraphics[width=\textwidth]{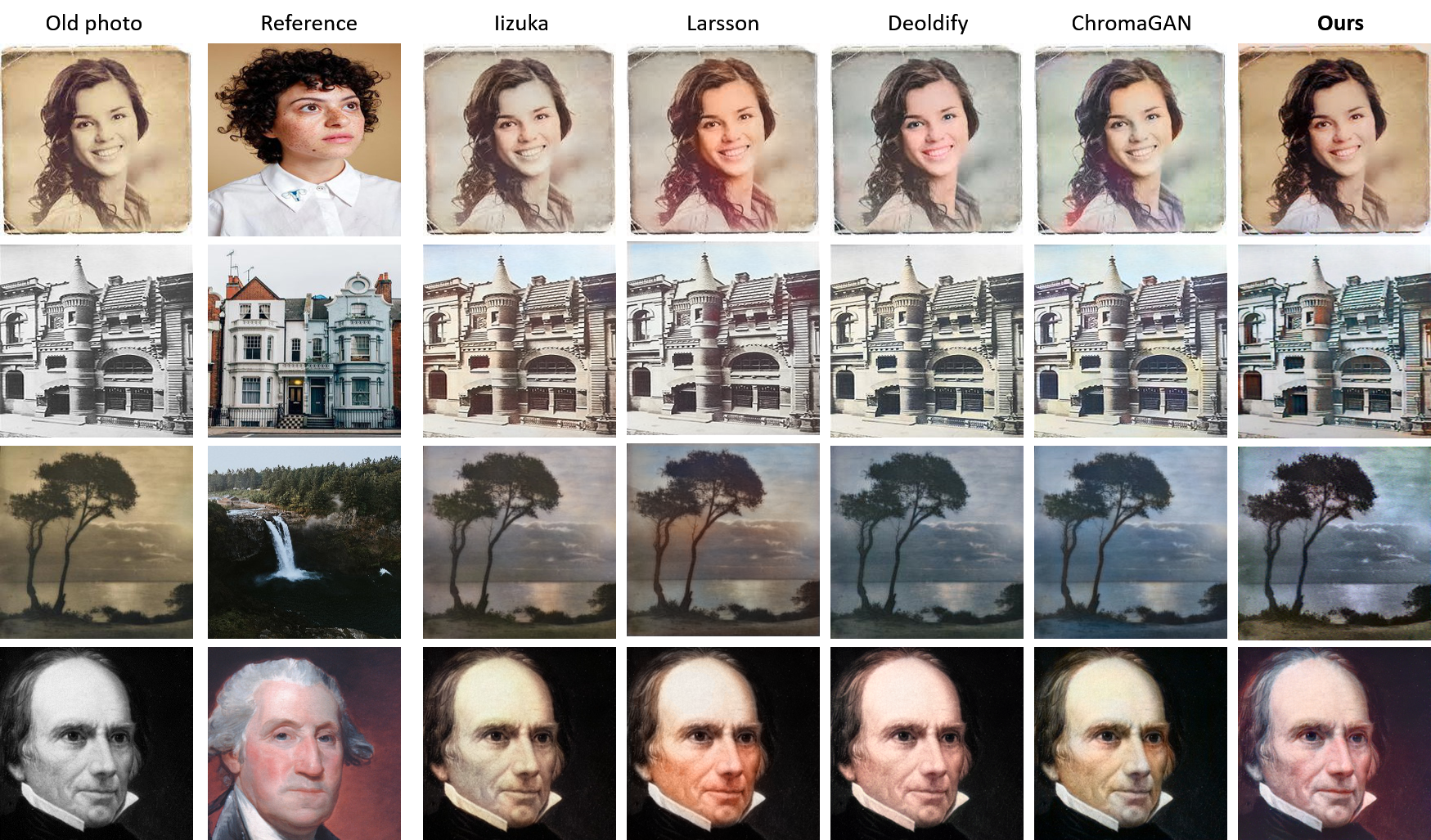}
	\caption{Comparisons with automatic colorization methods: Iizuka~\cite{iizuka2016let}, Larsson~\cite{larsson2016learning}, DeOldify~\cite{deoldify}, ChromaGAN~\cite{vitoria2020chromagan}. The reference images are only used in our method.}
	\label{fig:exper2}
\end{figure}
\noindent{\textbf{Comparison with Fully Automatic Colorization}.} 
Notably, the selected four methods are all trained on large-scale datasets with gray/color pairs. As shown in Fig~\ref{fig:exper2}, our method could generate more colorful results compared to all the baselines since we leverage reference images as color guidance. Our model can better achieve semantic-level color transfer, such as the colors of this woman's hair, face and clothes shown in row 1 in Fig.~\ref{fig:exper2}, whereas Iizuka~\cite{iizuka2016let}, Larsson~\cite{larsson2016learning}, DeOldify~\cite{deoldify} and ChromaGAN~\cite{vitoria2020chromagan} only mimic the global color statistics (\egno, row 2\&3 in Fig.~\ref{fig:exper2}) and generate strange color deviation (\egno, the soiled white of \cite{iizuka2016let}, the pale orange of \cite{larsson2016learning} and the green human face of \cite{vitoria2020chromagan}).

\begin{table}[ht]
	\centering
	\caption{FID and US scores toward fully automatic colorization methods. The best scores are in bold.}
	\begin{tabular}{c|ccccc}
		\toprule
		&  Iizuka~\cite{iizuka2016let} & Larsson~\cite{larsson2016learning} & DeOldify~\cite{deoldify}  & ChromaGAN~\cite{vitoria2020chromagan} &\textbf{Ours} \\
		\midrule
		FID & 246.91& 255.68 & 240.41 & 258.86 & \textbf{239.16} \\
		US & 24.50\% & 28.25\% & 44.75\% & 38.25\% & -\\
		\bottomrule
	\end{tabular}
	\label{tab:fully}
\end{table}

As for the quantitative evaluation, we measure the FID score only since this setting does not have corresponding reference images to calculate SIFID and CH scores. Similarly, we also conduct a user study with the usage of 40 groups of images containing various types of old photos. Each group contains one old photo and two outputs from another baseline method and ours. Then 20 participants are given unlimited time to answer the following question: Which colorized result do you prefer? We collect total $1600$ responses and $80$ responses for each comparison. As shown in Table.~\ref{tab:fully}, our method gets the lowest FID score and earns the most preferences on aesthetic traits.

\begin{figure}[!t]
	\centering
	\includegraphics[width=\textwidth]{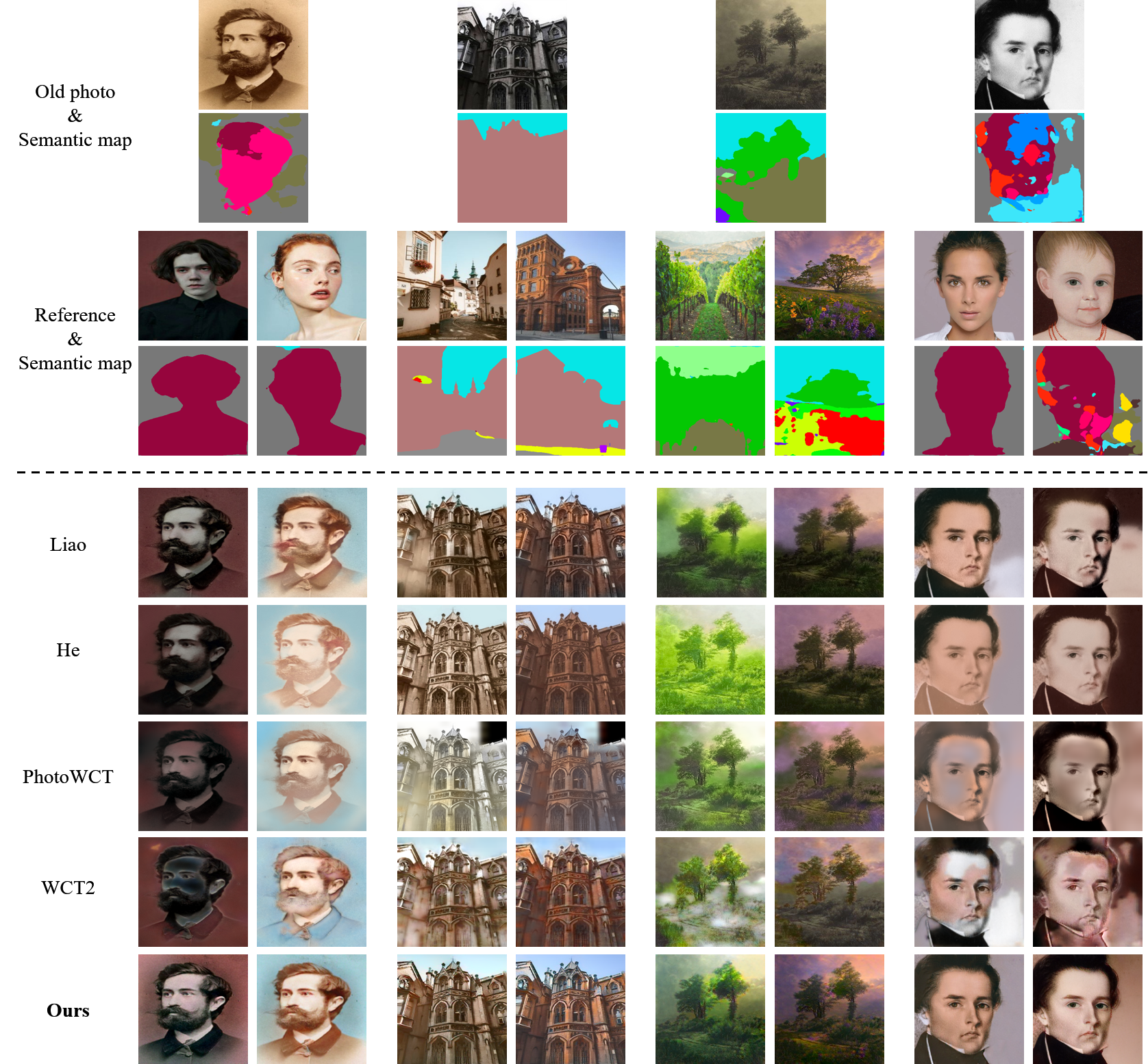}
	\caption{Comparisons with photorealistic style transfer methods: Liao~\cite{liao2017visual}, He~\cite{he2019progressive}, PhotoWCT~\cite{li2018closed}, WCT2~\cite{yoo2019photorealistic}.}
	\label{fig:exper3}
\end{figure}

\noindent{\textbf{Comparison with Photorealistic Style Transfer}.} 
Typically, style transfer involves two main types of tasks: artistic style transfer and photorealistic style transfer.
The primary distinction between artistic style transfer and old photo colorization is whether the structure is preserved exactly or not. Artistic style transfer does not guarantee a faithful representation of the original structure, often producing stylized results with randomly distributed textures. Old photo colorization requires that the original structure of the old photo must remain unchanged. Therefore, we compare with photorealistic style transfer which transfers the style of one high-quality natural image to another but preserves the original structure and detailed outline of the content image. 

The core difference between photorealistic style transfer and old photo colorization lies in how strictly the structure is preserved. Generally, existing photorealistic style transfer utilizes pre-trained CNNs to extract structural features which contain inseparable color information, thus interfering with color transfer and structure preservation (as demonstrated in Fig. ~\ref{fig:dino}). To address this, we replace the commonly used VGG with a vision transformer to capture structural information more effectively and ignore the effect of color. Furthermore, existing photorealistic style transfer necessitates either elaborate losses design \cite{liao2017visual, he2019progressive} or additional provision of semantic segmentation maps \cite{li2018closed, yoo2019photorealistic} for better structure and detail preservation. In contrast, our model does neither of these while producing more vivid and realistic colorization results; it only needs to freeze the non-high-frequency bands from participating in backpropagation during the iterative updating process.

As shown in Fig.~\ref{fig:exper3}, Liao~\cite{liao2017visual} generates results with awful spatial distortion and introduces unappealing color patterns because PatchMatch transfers visual attributes including color, texture, and strokes. He~\cite{he2019progressive} addresses the spatial distortion issue with the additional edge-preserving filter but it would over-smooth and blur the final result. 
Similarly, PhotoWCT~\cite{li2018closed} preserves local structures with smoothing loss to encourage pixel similarity resulting in noticeable artifacts and disharmonious color leaks. For WCT2~\cite{yoo2019photorealistic}, we provide coarse semantic maps generated by the SOTA segmentation algorithm~\cite{SunZJCXLMWLW19} to match its experimental setting. However, it still fails on nearly all scenes and generates results with worse semantic matching colors as well as obvious color drifting artifacts (\egno, the irregular brightness on faces and trees in column 1\&5\&7). In contrast, our method leverages FDA and SPM respectively to transfer color and preserve the structure, which produces more vivid and realistic colorization results.

\begin{table}[ht]
	\centering
	\caption{SIFID, FID, CH and US scores across different photorealistic style transfer methods. The best scores are in bold.}
	\resizebox{\textwidth}{!}{
		\begin{tabular}{c|ccccccccccccccccc}
			\toprule[1.5pt]
			\multirow{3}{*}{Configuration} & \multicolumn{4}{c}{\multirow{2}{*}{\begin{tabular}[c]{@{}c@{}}SIFID\\ (x10-4)\end{tabular}}} &  & FID &  & \multicolumn{8}{c}{CH} &  & US \\ \cline{7-7} \cline{9-16} \cline{18-18} 
			& \multicolumn{4}{c}{} &  & - &  & \multicolumn{4}{c|}{KL div.} & \multicolumn{4}{c}{H dis.} &  & - \\ \cline{2-5} \cline{9-16}
			& portrait & architectural & landscape & faces &  & - &  & portrait & architectural & landscape & \multicolumn{1}{c|}{faces} & portrait & architectural & landscape & faces &  & - \\ \cline{1-5} \cline{7-7} \cline{9-16} \cline{18-18} 
			Liao~\cite{liao2017visual} & 57.41 & 0.89 & 1.95 & 0.33 &  & 233.98 &  & \textbf{0.497} & 0.613 & \textbf{0.436} & \textbf{0.168} & \textbf{0.279} & 0.331 & \textbf{0.301} & \textbf{0.197} &  & 30.50\% \\
			He~\cite{he2019progressive} & 57.50 & 0.77 & 1.76 & 0.24 &  & 254.71 &  & 0.540 & 0.670 & 0.785 & 0.234 & 0.324 & 0.351 & 0.391 & 0.222 &  & 34.00\% \\
			PhotoWCT~\cite{li2018closed} & 57.49 & 0.76 & 1.79 & 0.34 &  & 262.37 &  & 1.031 & 1.832 & 1.007 & 0.611 & 0.478 & 0.536 & 0.419 & 0.378 &  & 20.50\% \\
			WCT2~\cite{yoo2019photorealistic} & 57.37 & 0.66 & 1.35 & 0.28 &  & 284.12 &  & 0.611 & 0.236 & 0.697 & 0.397 & 0.344 & 0.234 & 0.371 & 0.315 &  & 14.25\% \\
			\textbf{Ours} & \textbf{57.34} & \textbf{0.32} & \textbf{0.59} & \textbf{0.23} &  & \textbf{232.05} &  & 0.553 & \textbf{0.155} & 0.465 & 0.186 & 0.336 & \textbf{0.186} & 0.312 & 0.199 &  & \textbf{-}\\
			\bottomrule
	\end{tabular}}
	\label{tab:style}
\end{table}

As shown in Table.~\ref{tab:style}, we quantitatively evaluate the performance across different photorealistic style transfer methods on three objective metrics. One subjective user study is also conducted following the same experimental setup as the comparison of reference-based colorization. The lowest scores of SIFID and FID indicate that our model has the best visual effect. Liao~\etal\cite{liao2017visual} gets the lowest CH scores in most cases because they utilize PatchMatch to construct results with the visual details (\ieno, patches) directly sampled from the reference image. But their result suffers from obvious texture artifacts. What's more, we get the highest user preference because of the more faithful and saturated results.

\subsection{Ablation Study}

\begin{figure}[!t]
	\centering
	\includegraphics[width=1\textwidth]{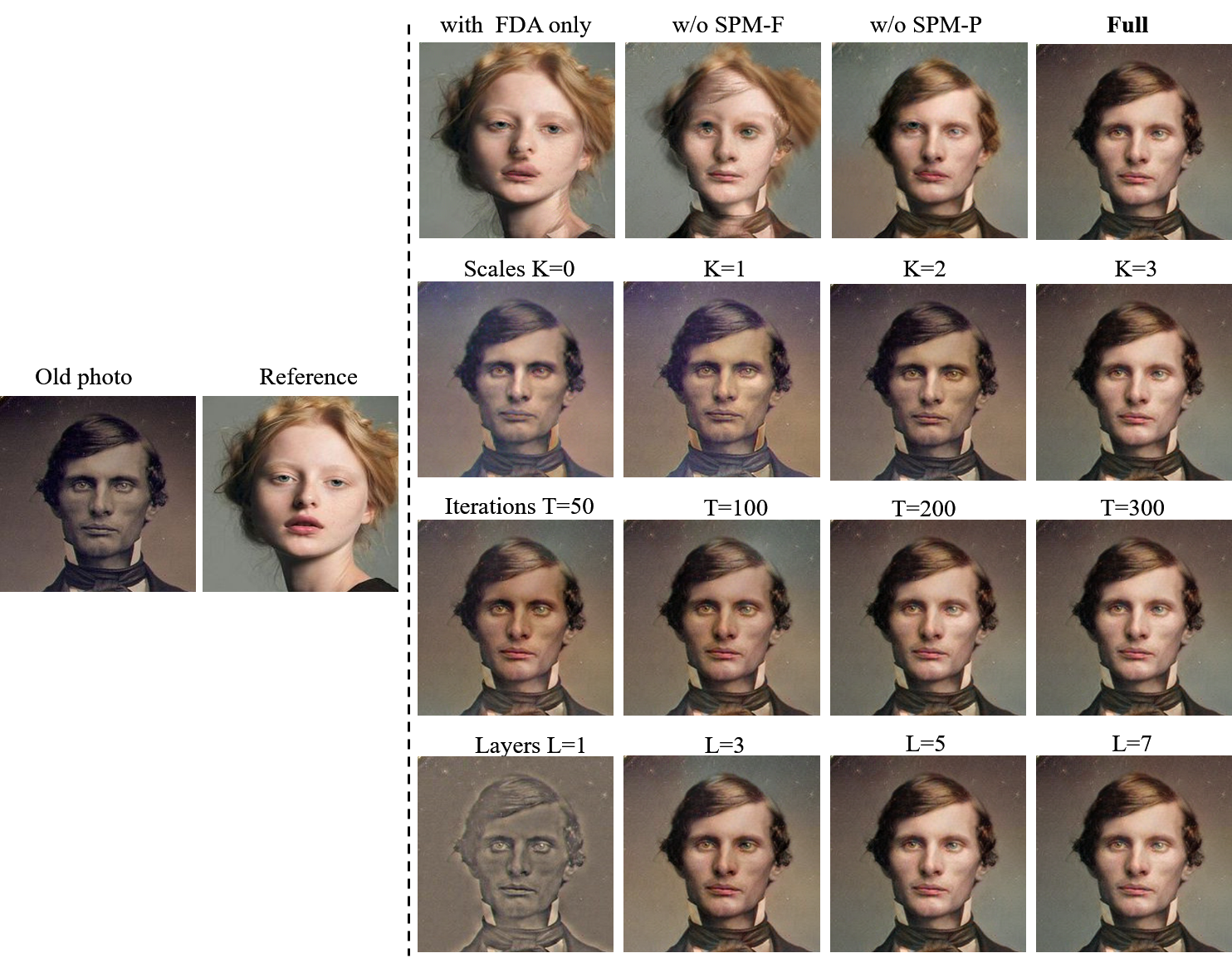}
	\caption{The first row indicates the ablation studies of different modules in our model, including optimizing using FDA only, without SPM-F (w/o SPM-F), without SPM-P (w/o SPM-P), and full modules (Full). We also examine the effect of number selection on three hyper-parameters: the scales of model $K$ in the second row, the iteration times $T$ in the third row, and the layers of the Laplacian pyramid $L$ in the last row. }
	\label{fig:ab1}
\end{figure}

\begin{table}[ht]
	\centering
	\caption{Ablation study on the influence of different components and the selection of scale numbers $K$, iteration times $T$ and Laplacian pyramid layers $L$. The best scores are in bold.}
	\resizebox{\textwidth}{!}{
		\begin{tabular}{c|c|cccccccccccccccccccc}
			\toprule
			\multicolumn{2}{c|}{Metric} &  & with FDA only & w/o SPM-F & w/o SPM-P & Full &  & K=0 & K=1 & K=2 & K=3 &  & T=50 & T=100 & T=200 & T=300 &  & L=1 & L=3 & L=5 & L=7 \\ \cline{1-2} \cline{4-7} \cline{9-12} \cline{14-17} \cline{19-22} 
			\multicolumn{2}{c|}{SIFID($\times10^{-5}$)} &  & \textbf{3.04} & 7.38 & 5.86 & 7.07 &  & 23.04 & 17.95 & 13.84 & \textbf{7.07} &  & 9.50 & 8.44 & 7.63 & \textbf{7.07} &  & 29.31 & 7.20 & 7.07 & \textbf{6.57} \\ \cline{1-2}
			\multicolumn{2}{c|}{FID} &  & 263.83 & 274.20 &235.20 &\textbf{223.67} &  &240.15 & 249.89 & 234.91 & \textbf{223.67} & & 244.33 & 241.30 & 236.01 & \textbf{223.67} & & 354.59 & 238.93 & 223.67 &\textbf{219.69}    \\ \cline{1-2}
			\multirow{2}{*}{CH} & KL Div. &  & \textbf{0.083} & 0.101 & 0.156 & 0.307 &  & 0.462 & 0.469 & 0.396 & \textbf{0.307} &  & 0.384 & 0.366 & 0.340 & \textbf{0.307} &  & 0.538 & 0.292 & 0.307 & \textbf{0.261} \\ \cline{2-2}
			& H Dis. &  & \textbf{0.028} & 0.042 & 0.098 & 0.462 &  & 1.021 & 1.109 & 0.912 & \textbf{0.462} &  & 0.387 & 0.264 & 0.613 & \textbf{0.462} &  & 1.690 & 0.373 & 0.462 & \textbf{0.312} \\
			\bottomrule
	\end{tabular}}
	\label{tab:abl}
\end{table}

\noindent{\textbf{Influence of Components}.} We study the influence of different modules in our model, including optimizing using FDA only, without SPM-F (w/o SPM-F), without SPM-P (w/o SPM-P), and full modules (Full). The ablation results in Fig.~\ref{fig:ab1} show that training with FDA only on two images would force the model to align the feature distribution of the result to the reference by directly copying the reference image since there is no extra constraint. Without SPM-F, the model will introduce obvious content distortions. Without SPM-P, the model cannot preserve edge details and generates weird spatial distortion (\egno, the left eyes). Not only our full model can transfer the color from reference as much as possible but also maintain raw content structure. The quantified results are shown in Table.~\ref{tab:abl}, Full obtains the best FID score, which indicates that our SFAC can generate more realistic and vivid colorized photos. Besides, with FDA only gets the best SIFID and CH scores because it directly copies the reference image.

\noindent{\textbf{Selection of Numbers}.}
We also examine the effect of the selection of numbers on three hyper-parameters: the scales of model $K$, the iteration times $T$, and the layers of the Laplacian pyramid $L$. The effect is illustrated in Fig.~\ref{fig:ab1}. As can be seen, as the number of scales $K$ increases, colors transferred from the reference are more abundant. As for iterations $T$ and layers $L$, the number is larger and the performance is better. When the number is large enough, they get convergence gradually. In most cases, we set $K=3$, $T=200$, and $L=5$ considering the trade-off between algorithm complexity and colorization performance. As shown in Table.~\ref{tab:abl}, $K=3$, $T=300$ and $L=7$ all get the best scores, and setting $T=300$ and $L=7$ indeed improves performance slightly, but increases time consuming.


%
\begin{figure}[ht]
	\centering
	\includegraphics[width=0.8\textwidth]{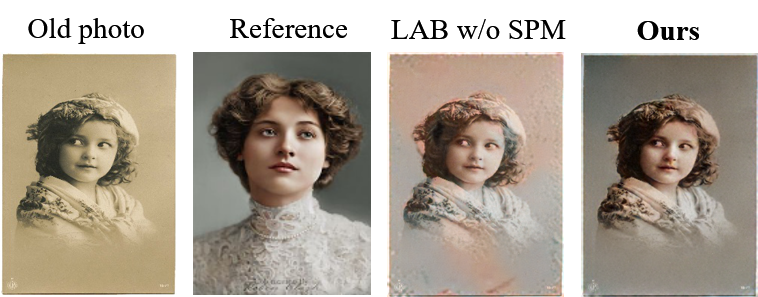}
	\caption{The comparison between optimizing ab channels only without SPM (LAB w/o SPM) and optimizing rgb channels with SPM (Ours).}
	\label{fig:lab}
\end{figure}
\noindent{\textbf{Comparison between RGB and LAB}.
Reference-based image colorization~\cite{10.1145/3197517.3201365, zhang2019deep, xiao2020example} first obtain inaccurate warped ab\footnote{Colorization task defining images on CLELab color space typically utilizes gray images as L* channel to predict corresponding a*b* values.} values based on the semantic correspondence between the reference image and the old photo, and then randomly sample them as inputs for the learning-based colorizer to propagate complete ab information. With the help of extensive training data, the colorizer can eventually equate the semantic correspondence of features with the color correspondence of ab values. However, we cannot achieve such color warping with only two training images, the unreliable warped ab values would seriously hamper the color transfer. Therefore, we learn the color warping directly by feature distribution alignment with the hierarchical features defined on the RGB color space. Moreover, colorizing old photos in the RGB color space can acquire more enhanced and vivid results with both luminance and chrominance faithful to the reference image, which cannot be achieved by those existing colorization methods that must preserve the original luminance and optimize ab color channels. To better verify the superiority of our method, we conduct the comparative experiment to convert images to LAB space and only optimize a*b* channels without SPM (LAB w/o SPM). As shown in Fig.~\ref{fig:lab}, LAB w/o SPM gets awful colorized results, such as the color mixture issue of hair and background, even though it leverages the semantic correspondence of features, while our method yields more saturated and vibrant results.

\begin{figure}[ht]
	\centering
	\includegraphics[width=\textwidth]{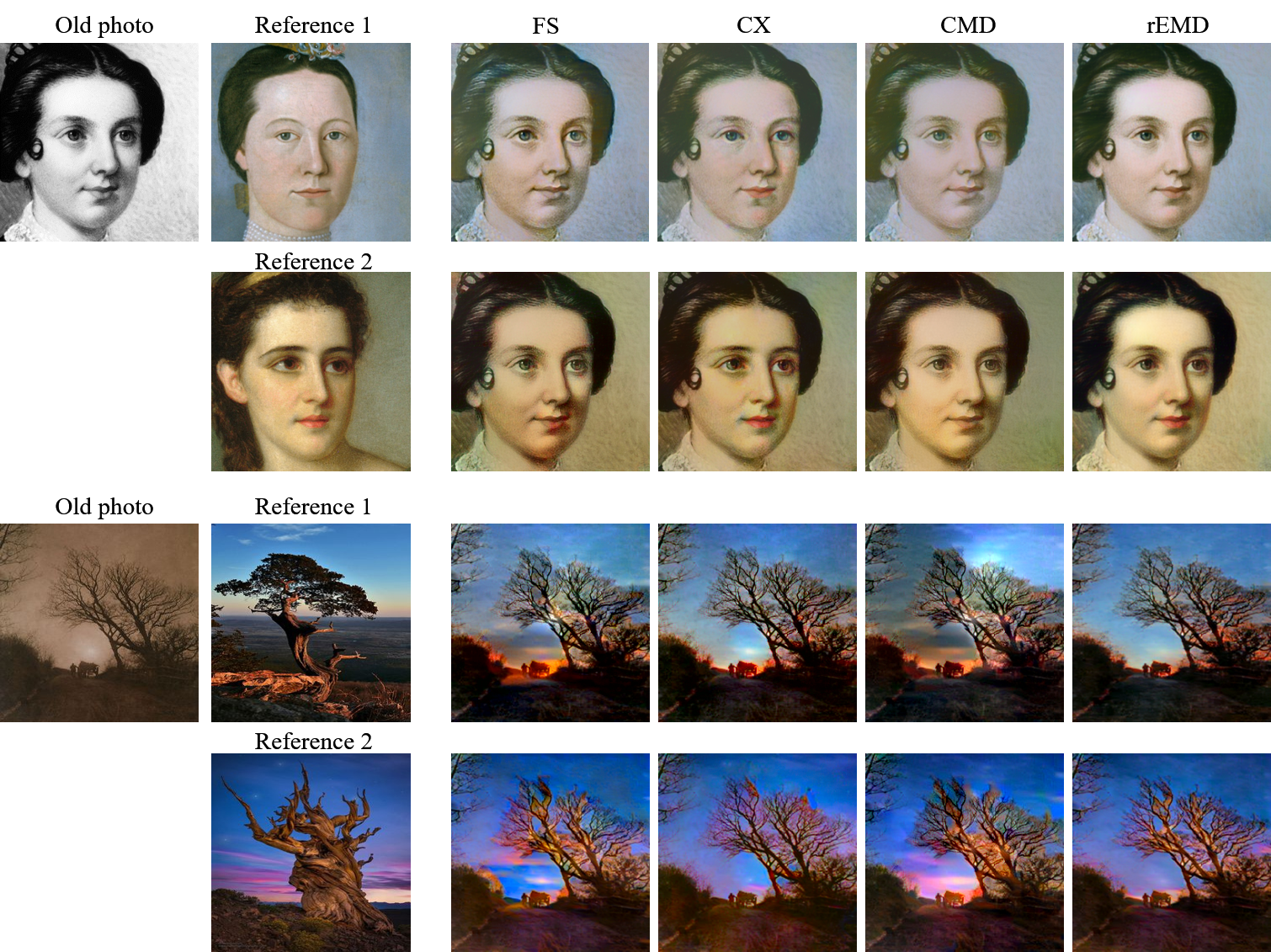}
	\caption{The colorization results when using different metrics in FDA.}
	\label{fig:FDA}
\end{figure}

\noindent{\textbf{Effectiveness of FDA}.}
To better verify the effectiveness of our proposed FDA for colorization, we choose different metrics to measure the distance. As shown in Fig. \ref{fig:FDA}, all metrics can successfully colorize the old photo even though these results may be visually different (\egno, the sunset over the wagon). It also reveals that our method is robust, and one can design different distance metrics to further improve the performance under the FDA framework. 

\begin{figure}[ht]
	\centering
	\includegraphics[width=\textwidth]{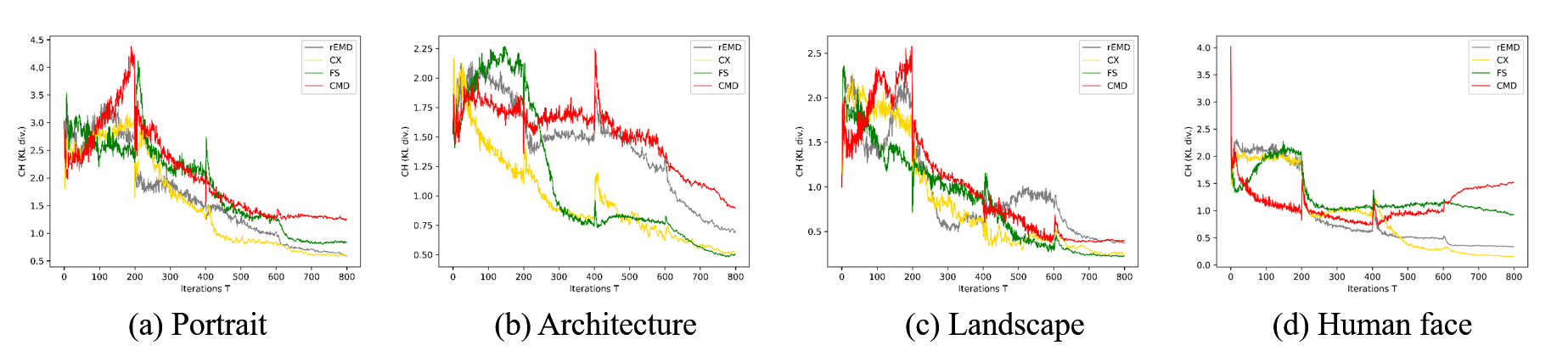}
	\caption{Convergence curves of four metrics on different types of old photos by measuring the CH difference (KL div.) between the colorized photo and its reference with iterations $T$.}
	\label{fig:metrics}
\end{figure}
\noindent{\textbf{Analysis of metrics used in FDA}.}
As shown in Fig.~\ref{fig:metrics}, we also focus on the detailed distinctions of the four metrics. For better understanding, we visualize their CH (KL Div.) convergence curves with the number of iterations $T$ to illustrate the color transfer performance on different data types. In general, we set CX as our default metric due to its robustness and superiority in most cases. FS performs better in architectural and landscape photography because it only matches the mean and variance of the feature distribution. For images with regular homogeneous color regions and no clear color boundaries (\egno, house and sky), these two statistics convey sufficient color information. rEMD is particularly good for portraiture and human faces as it takes into account the geometry of the underlying space, ensuring one-to-one mapping (\egno, eye colors and lipsticks in Fig.~\ref{fig:FDA}). CMD presents a more unstable trait (\egno, the growing trend of the final scale in Fig.~\ref{fig:metrics} (d)) since it aligns the two distributions with five moments, and changes in higher order statistics containing high-frequency structural information would be penalized by SPM.

\section{Conclusion}
In this paper, we propose a CNN-based algorithm for old photo colorization, explicitly trained on only two images to avoid extensive data and domain gaps. We introduce the FDA loss to simplify the process of establishing semantic correspondence and color transfer. Our FDA exhibits robustness to various distribution distance metrics. To preserve content, we enforce a perceptual constraint using transformer-based deep features, effectively mitigating the impact of color effects at the feature level. At the pixel level, we employ a frozen-updated pyramid design to retain details and boundaries. Extensive experimental results demonstrate the wide applicability of our model across different scenes, yielding vibrant and saturated colorized results.

\begin{figure}[!t]
	\centering
	\includegraphics[width=\textwidth]{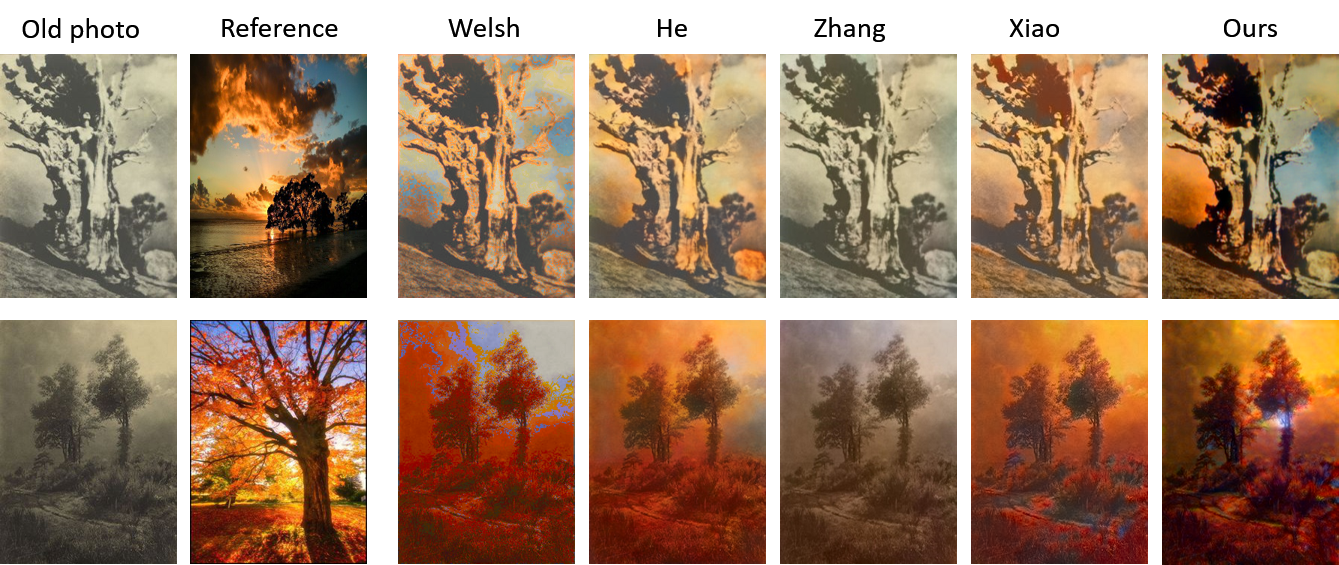}
	\caption{The failed cases: Welsh~\cite{welsh2002}, He~\cite{10.1145/3197517.3201365}, Zhang~\cite{zhang2019deep}, Xiao~\cite{xiao2020example}.}
	\label{fig:failed}
\end{figure}

However, the method has limitations, as shown in Fig.~\ref{fig:failed}. SFAC underperforms in landscape photography due to several reasons. Firstly, the presence of various deteriorations in low-quality old photos, such as noise, scratches, and low contrast, hampers accurate recognition. Secondly, the lack of clear contour boundaries in chaotic scenes presents challenges for FDA in establishing appropriate semantic correspondence. And the experimental results also show the superior performance of our model in old photographs with higher quality and clearer structural elements (see the supplemental material). To address this issue, we propose exploring the use of additional segmentation maps to guide colorization as a potential solution, which will be investigated in our future research.



 \bibliographystyle{elsarticle-num} 
 \bibliography{main}





\end{document}